\documentclass[10pt,twocolumn,letterpaper]{article}

\usepackage{iccv}
\usepackage{times}
\usepackage{epsfig}
\usepackage{graphicx}
\usepackage{amsmath}
\usepackage{amssymb}
\usepackage{microtype}      
\usepackage{subcaption}
\usepackage{bm}
\usepackage{multirow}
\usepackage{subcaption}
\usepackage{sidecap}
\usepackage{url}            
\usepackage{booktabs}       
\usepackage{amsfonts}       
\usepackage{nicefrac}       
\usepackage{pifont}
\newcommand{\xmark}{\ding{55}}%

\usepackage[pagebackref=true,breaklinks=true,letterpaper=true,colorlinks,bookmarks=false]{hyperref}

\iccvfinalcopy 

\def\iccvPaperID{3} 

\ificcvfinal\fi

\begin{document}

\title{Self-supervised Visual Attribute Learning for Fashion Compatibility}

\author{Donghyun Kim, Kuniaki Saito, Samarth Mishra, Stan Sclaroff, Kate Saenko, Bryan A. Plummer \\
Boston University \\
\tt\small \{donhk, keisaito, samarthm, sclaroff, saenko, bplum\}@bu.edu
}

\maketitle
\ificcvfinal\thispagestyle{empty}\fi

\begin{abstract}
   Many self-supervised learning (SSL) methods have been successful in learning semantically meaningful visual representations by solving pretext tasks. However, prior work in SSL focuses on tasks like object recognition or detection, which aim to learn object shapes and assume that the features should be invariant to concepts like colors and textures. Thus, these SSL methods perform poorly on downstream tasks where these concepts provide critical information. In this paper, we present an SSL framework that enables us to learn color and texture-aware features without requiring any labels during training.  Our approach consists of three self-supervised tasks designed to capture different concepts that are neglected in prior work that we can select from depending on the needs of our downstream tasks. Our tasks include learning to predict color histograms and discriminate shapeless local patches and textures from each instance. We evaluate our approach on fashion compatibility using Polyvore Outfits and In-Shop Clothing Retrieval using Deepfashion, improving upon prior SSL methods by 9.5-16\%, and even outperforming some supervised approaches on Polyvore Outfits despite using no labels.  We also show that our approach can be used for transfer learning, demonstrating that we can train on one dataset while achieving high performance on a different dataset.
\end{abstract}

\begin{figure*}[t]
	\centering
    \begin{subfigure}[t]{0.3\textwidth}
        \includegraphics[width=5cm, height=5cm]{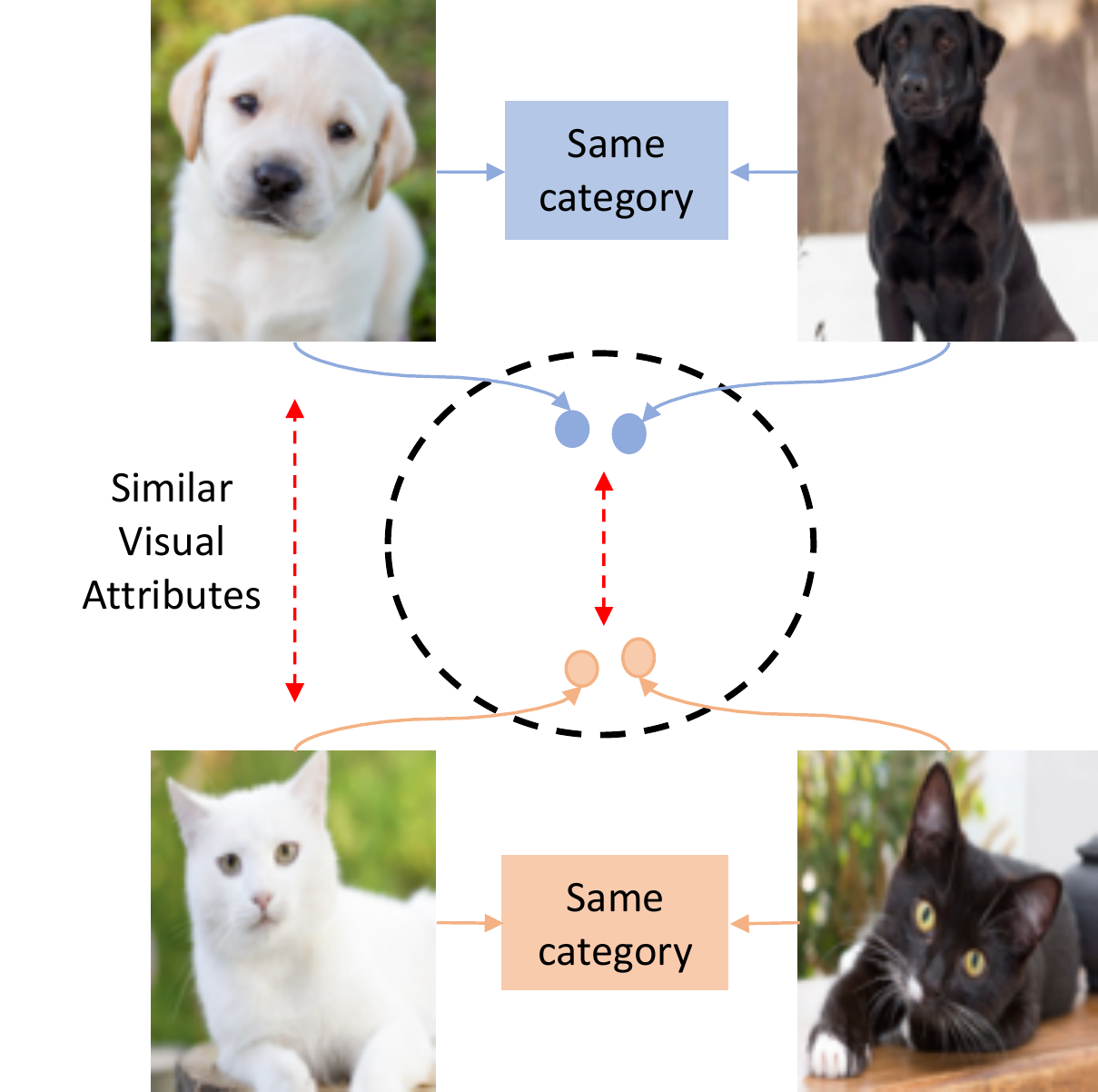}
        \caption{Object Recognition}
    \end{subfigure}
    ~
	\begin{subfigure}[t]{0.3\textwidth}
        \centering
        \includegraphics[width=1.0\textwidth, height=5cm]{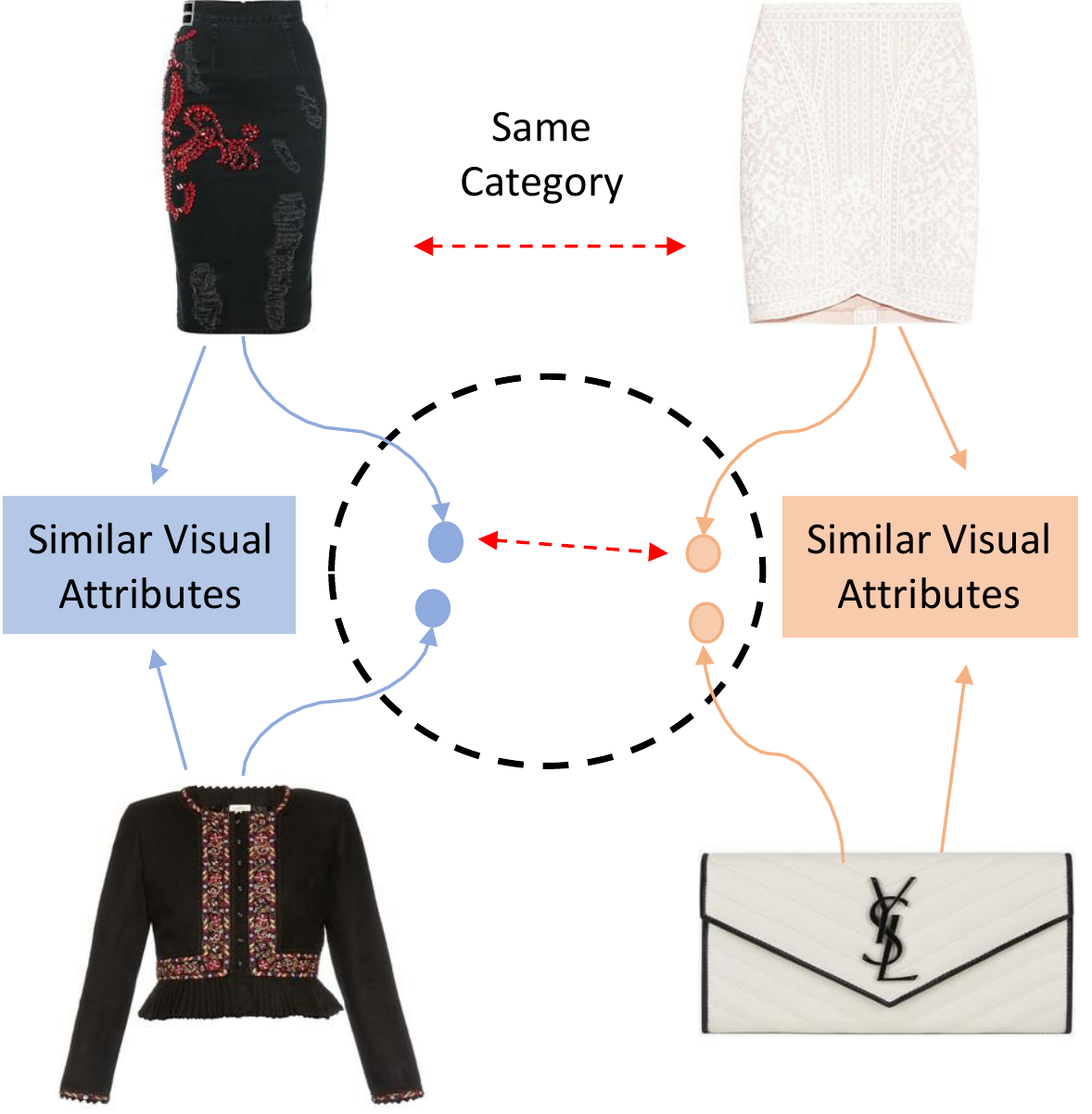}
        \caption{Fashion Compatibility}
    \end{subfigure}
    ~
    \begin{subfigure}[t]{0.3\textwidth}
        \centering
        \includegraphics[width=0.7\textwidth, height=5cm]{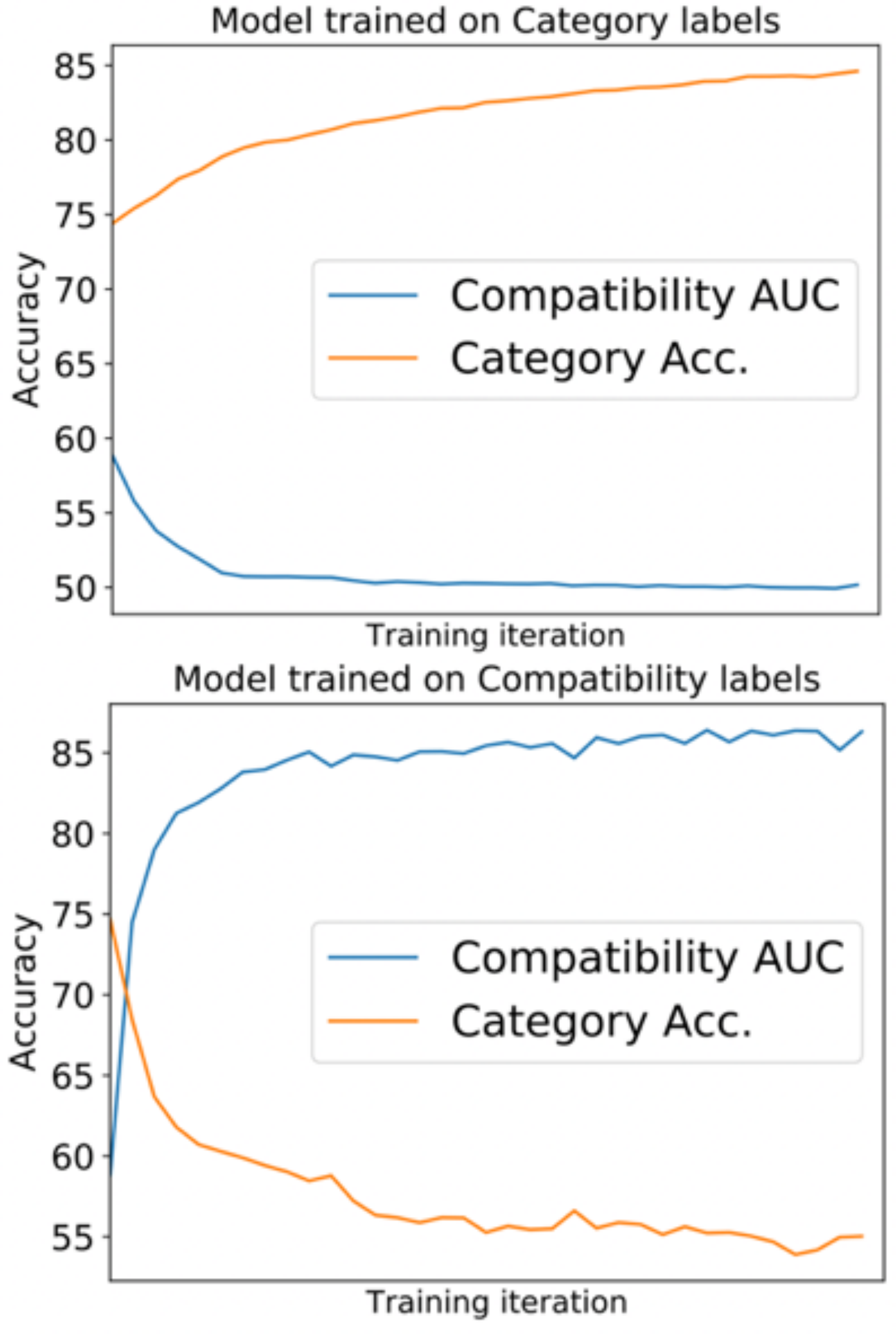}
        \caption{Non-transferability}
    \end{subfigure}
	\caption{Differences between the (a) object recognition and (b) fashion compatibility tasks. (a) Object recognition needs \textit{color invariant} but \textit{shape sensitive} features. (b) Tasks like fashion compatibility needs \textit{color sensitive} but \textit{shape invariant} features in order to match different category fashion items, in which items of the same object category can be embedded far under different visual attributes. In (c), we show that a model trained on object category labels hurts performance on the fashion compatibility task and vice versa, which helps motivate us to propose a new form of SSL pretext tasks.
	} 
	\label{fig:fig_intro}
\end{figure*}

\section{Introduction}
Colors and textures information are important features for tasks like fine-grained classification~\cite{attributes2012cvpr,Gebru_2017_ICCV} and 
microscopy image classification~\cite{micosopy} as well as applications like image search, recommendation, and outfit generation~\cite{cucurull2019context,han2017learning, hsiao2018creating, tan2019learning,FedCSIS201756,vasileva2018learning,veit2015learning}.  However, collecting annotations to train these models can be expensive, especially when they require domain expertise~\cite{perone2019unsupervised} or are constantly evolving like e-commerce datasets. Self-supervised learning (SSL) would appear to be a good fit to address this problem since they require no labels for training, but prior work focused on tasks like object classification and detection (\eg~\cite{gidaris2018unsupervised,noroozi2016unsupervised,wu2018unsupervised,chen2020simple,he2019momentum}), where the goal is to recognize an object (\ie, its shape) regardless of its color or texture (so \textit{a black dog} and \textit{a white dog} should both be classified as \textit{a dog}).  In fact, many self-supervised approaches are explicitly designed to learn color invariant features~\cite{chen2020simple,he2019momentum}.  Thus, as we illustrate in Figure~\ref{fig:fig_intro}, prior work in SSL often does not generalize to tasks where colors and textures are important.

\begin{figure*}[t!]
	\centering
    \includegraphics[width=0.75\linewidth]{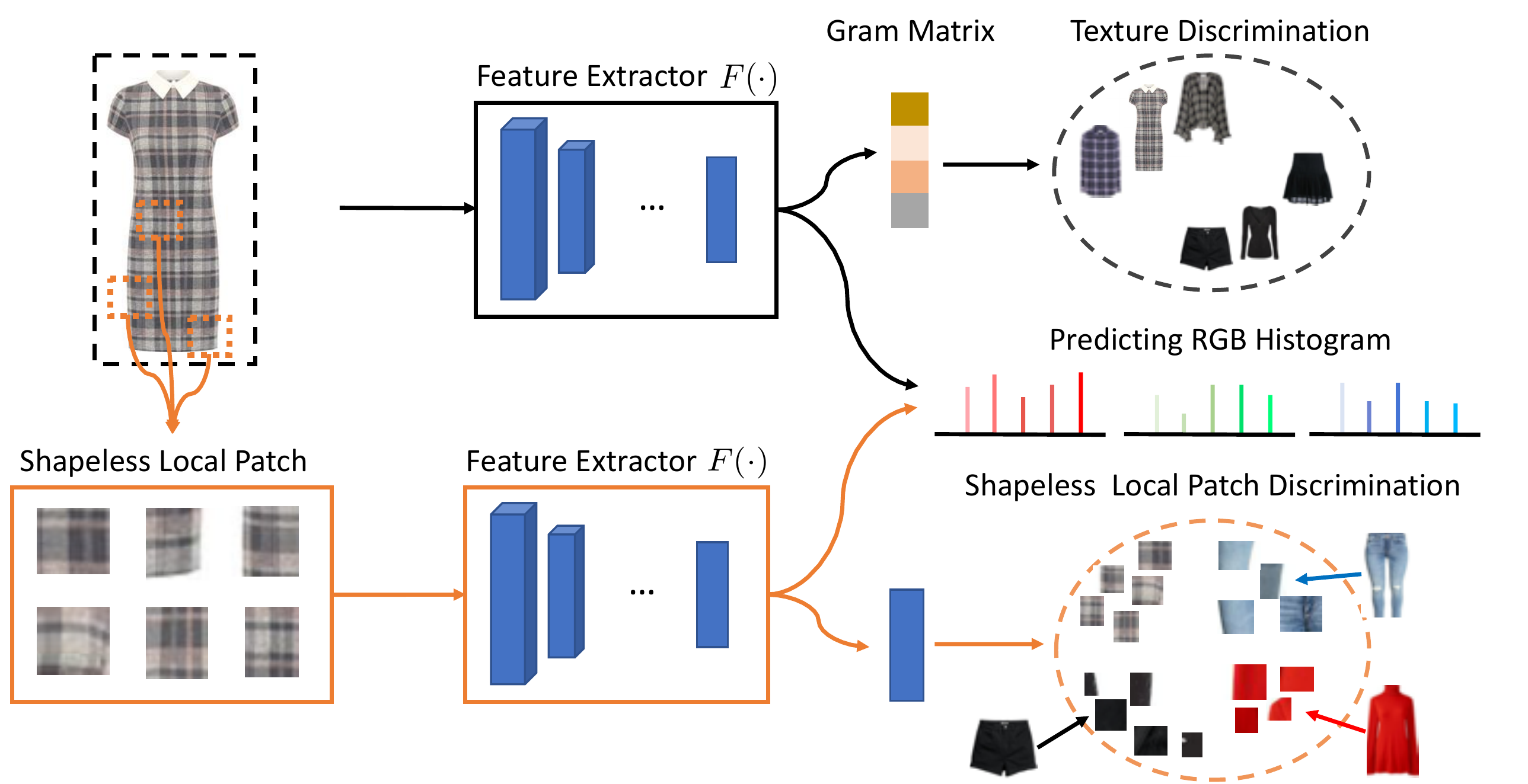}
	\caption{An overview of our Self-supervised Tasks for Visual Attribute (S-VAL), where we aim to learn discriminative features in colors and textures without encoding shape information. To achieve this goal, we propose thee sub-tasks (1) predicting RGB histogram, (2) shapeless local patch discrimination, and (3) texture discrimination. 
    }
	\label{fig:fig_method}
\end{figure*}

In this paper, we propose Self-supervised Tasks for Visual Attribute (S-VAL) to learn visual attributes while generating \textit{shape invariant} features for fashion compatibility, where a system recommends fashion items compatible and complement each other when worn together in an outfit.  Motivated by the observation that similar color or texture items are likely to be compatible~\cite{plummerSimilarityExplanations2020}, S-VAL is designed to learn embedding images with similar colors and texture patterns are embedded nearby each other. To be specific, our approach consists of three major components.  First, we propose a new self-supervised pretext task where a model predicts color histograms of input images to understand dominant colors of an image. Second, we introduce shapeless local patch discrimination, where we perform Instance Discrimination (ID)~\cite{wu2018unsupervised} on very small image patches of an image. This helps ensure that little shape information is present in an image and the model must focus on recognizing color and texture information instead.  Finally, we obtain texture features using a Gram matrix~\cite{gatys2015neural,lin2015bilinear,lin2016visualizing} computed over the whole image, and then encourage ID to learn discriminative texture representations. Our approach uses no labels in training (\ie, unsupervised), but, as our experiments will show, we get comparable performance to some fully-supervised methods.  Figure~\ref{fig:fig_method} provides an overview of our approach.

The work that is the closest in spirit to ours is Hsiao~\etal~\cite{hsiao2018creating}, which automatically identifies individual clothing items from full-body photos of people and then uses the parsed outfits as labels for fashion compatibility.  This is reminiscent of the part-based methods used in tasks like object classification~\cite{Endres_2013_CVPR}, where the goal is to learn how to identify the parts (or individual clothing items) in order to recognize the object (or to recognize compatible items).  However, this still requires having weak-labels and is a task-specific application (\ie, it only is applicable to fashion compatibility).  Another significant drawback is that the images used for training were from a different domain (full-body images of people) than the images they are evaluated on (images containing a single product on a white background). Thus, as our experiments will show, our approach significantly outperforms the weakly-supervised approach of Hsiao~\etal~\cite{hsiao2018creating} despite our approach lacking any supervision and without making task-specific assumptions. Specifically, in addition to comparing to Hsiao~\etal on fashion compatibility, we also evaluate our approach on In-Shop Clothing Retrieval~\cite{liu2016deepfashion}, demonstrating that our approach generalizes.

Our contributions are summarized below:
\vspace{-2mm}
\begin{itemize}
\setlength\itemsep{0em}
    \item We propose Self-supervised Tasks for Visual Attribute (S-VAL) to learn \textit{colors and textures of images} while generating \textit{shape invariant} features. To the best of our knowledge, ours is the first work to propose SSL methods for capturing color and texture information.
    \item   We obtain a 9.5-16\% gain in fill-in-the-blank outfit completion using Polyvore Outfits~\cite{vasileva2018learning} and on In-Shop Retrieval using DeepFashion~\cite{liu2016deepfashion} over prior SSL methods. Notably, our approach outperforms some supervised methods on Polyvore Outfits despite using no labels.
    \item We show our approach creates powerful features that transfer across datasets.  Specifically, we train on Polyvore Outfits and test on Capsule Wardrobes~\cite{hsiao2018creating}, and train on the Fashion-Gen dataset~\cite{rostamzadeh2018fashion} and test on Polyvore Outfits, reporting a 6-8\% gain over prior work.
    \item We demonstrate that self-supervised learning should consider \textit{different characteristics of downstream tasks} by highlighting the difference between object recognition and tasks like fashion compatibility and image retrieval, which we hope inspires future work in SSL.
\end{itemize}

\section{Related Work}

\noindent\textbf{Self-supervised Learning (SSL).} Self-supervised learning~\cite{gidaris2018unsupervised,noroozi2016unsupervised,wu2018unsupervised,he2019momentum,chen2020simple,misra2019self} generates self-supervisory signals for a pretext task from an input. By solving a pretext task, a model can learn semantically meaningful features from raw data. Handcrafted pretext tasks such as predicting rotations~\cite{gidaris2018unsupervised} and solving jigsaw puzzles~\cite{noroozi2016unsupervised} provide useful features for object recognition and detection tasks.  Wu~\etal~\cite{wu2018unsupervised} proposes an Instance Discrimination (ID) pretext task with contrastive loss~\cite{hadsell2006dimensionality}. ID learns visual similarity in different images by treating an image as its own class (\ie, positive pair) but all other images as negative pairs. While ID is effective at learning strong visual representations, ID can be biased to texture or colors of an object which is harmful to objection recognition. In later work, ID with strong data augmentation techniques like color distortion (\eg, color jittering and gray-scale images)~\cite{chen2020simple,chen2020improved} significantly improved the recognition or detection performance by providing color and texture invariant features. While these perform well for a task like object recognition or detection, they focus on learning an object's shape. For example, the images of ``black dog'' and ``white dog'' should be classified as the same class ``dog'' as shown in Fig.~\ref{fig:fig_intro}(a), so that ``black'' and ``white'' attribute should ideally be ignored. However, many tasks require reasoning about multiple similarity notions such as color, texture and style these methods ignore. Specifically, in tasks like fashion compatibility, where two items are considered compatible if they would complement each other when worn in the same outfit, items of different categories (\eg, a shirt and pants) can be compatible with each other. Thus, learning object categories can be harmful to performance (Fig.~\ref{fig:fig_intro}(c)). Instead, we propose an SSL framework that learns visual attributes for tasks where color and texture are important.
\smallskip

\noindent\textbf{Fashion Compatibility.} Other than the weakly-supervised approach of Hsiao~\etal~\cite{hsiao2018creating} we discussed in the Introduction, much of the recent work on fashion compatibility has assumed labels are available during training~\cite{cucurull2019context,han2017learning,vasileva2018learning,yang2019interpretable,tan2019learning,LIN_2020_CVPR,veit2015learning}.  Many of these approaches aim to decompose the fashion compatibility task into similarity conditions that may be learned automatically~\cite{tan2019learning,LIN_2020_CVPR} or could be explicitly defined~\cite{vasileva2018learning,yang2019interpretable,veit2017conditional}. All of these methods require many labels of positive pairs and arbitrarily choose negative samples, since datasets are not annotated with incompatible items, which can result in poor constraints~\cite{wu2017sampling}. Also, as our experiments will show, we outperform some supervised fashion compatibility methods without using any supervision.
\smallskip

\noindent\textbf{Visual Attribute Learning.} Visual attributes such as colors (\eg,~red, blue), texture  (\eg,~palm, colorblock), or fabric (\eg,~leather, tweed) provide natural visual patterns of fashion items. In order to learn these visual attributes in items, some methods~\cite{vittayakorn2016automatic,berg2010automatic} 
leverage visual attribute labels such as color or style extracted from text descriptions. However, these attribute labels can be very sparse and highly non-curated. Plummer~\etal~\cite{plummerSimilarityExplanations2020} introduce an attribute explanation model to find salient attributes for fashion item matching and find that colors are the one of the most salient attributes. Our SSL learns colors of fashion items and embed them near each other to build better representations for the task of fashion compatibility.

\section{S-VAL: Self-supervised Tasks for Visual Attribute Learning}
\label{sec:method}
We explore image similarity learning under an unsupervised setting where we have only unlabeled images $\mathcal{D}=\left\{\left(\mathbf{x}_{i}\right)\right\}_{i=1}^{N}$. These items include items of different categories such as pants, tops, and shoes. Compared to prior work in self-supervised learning (SSL), our approach aims to learn visual attributes without encoding any shape clues which could hurt downstream task performance (\ie,~shape-invariant features). Our SSL approach consists of three sub-tasks: (1) predicting color histograms, (2) shapeless local patch discrimination (SLPD), and (3) texture discrimination (TD). We train a model with three sub-tasks jointly. Our model consists of a CNN feature extractor $F(\cdot) \in \mathbb{R}^n$ and separate projection heads $C(\cdot)$ for each sub-tasks. Figure~\ref{fig:fig_method} contains an overview of our method.

\subsection{Predicting Color Histogram}
\label{sec:rgb}
Colors are salient attributes in tasks fashion compatibility~\cite{plummerSimilarityExplanations2020,tangseng2020toward,yang2019interpretable} or microscopy image classification~\cite{micosopy}. Thus, a color histogram of an item can provide useful properties of an image including its colors, contrast, and brightness of an item. 
In contrast to previous color reconstruction methods such as AutoEncoders~\cite{hinton1994autoencoders}, we learn to predict an RGB color histogram, which is an \textit{orderless} visual representation and therefore does not encode shape information~\cite{liu2019bow}. This means that objects from different categories (\eg, black top and black pants) can be embedded closely in the color embedding space. Given an image $\mathbf{x}$ with width $w$ and height $h$, we first compute the normalized histogram of $n$ bins for each $R,G, \text{and} B$ channels, for example,
\begin{equation}
h_r(l) = \frac{|\{i,j\}: e_{l} \leq \mathbf{x_r}(i,j)< e_{l+1}|}{w\times h}
\end{equation}
where $h_r$ represents the histogram of the $R$ channel of the image (\ie, $\mathbf{x_r}$) and $e_{l}$ is the $l$-th bin edge. $h_g$ and $h_b$ are defined similarly. In the case we are learning a presentation for product images commonly found in e-commerce websites, we exclude any white background pixel values.

From the image feature from a CNN (\ie, $\mathbf{f}=F(\mathbf{x})$), we compute predictions of histograms  for the $R$ channel $C_r(f) \in\mathbb{R}^n$, $G$  channel $C_g(f)\in\mathbb{R}^n$, and $B$ channel $C_b(f)\in\mathbb{R}^n$. In order to obtain the probability distributions of each channel (\ie~ $p_r, p_g, \text{and, }p_c$), we apply the softmax function. Then, we minimize the KL divergence between predicted distribution and the ground-truth histogram,
\vspace{-4mm}
\begin{equation}
\mathcal{L}_{rgb} = D_{KL} \left[ p_r \| h_r \right] + D_{KL} \left[ p_g \| h_g \right] + D_{KL} \left[ p_b \| h_b \right]
\end{equation}

\subsection{Shapeless Local Patch Discrimination (SLPD)}
\label{sec:slpd}
 While predicting histogram captures the dominant colors in images, it lacks in learning detailed color patterns such as the spatial organization of colors and textons in fashion items. In this section, we aim to learn discriminative color or texture representations by using shapeless local patches. In previous SSL methods, strong augmentation techniques with color distortion with Instance Discrimination (ID)~\cite{wu2018unsupervised,chen2020simple,chen2020improved} can be used together to become invariant to color or texture information so they learn to better identify shapes.  While this may be appropriate for tasks like object recognition, as shown in Fig.~\ref{fig:fig_intro}(c), learning shape information harms performance on tasks like fashion compatibility where image similarity is not determined completely by an item's shape.
 
To avoid focusing on shape, we perform ID on shapeless small local patches (SLP) that contain little or no shape information. Figure~\ref{fig:fig_method} shows examples of the SLPs. While random cropping has been used in prior work~\cite{chen2020simple,wu2018unsupervised}, they often use relatively large cropping ratios $r$ (\ie, [0.2, 1.0]) to maximize the consensus between local-to-global views.  However, these will often contain shape information, whereas SLP use very small ratio values of $r$ (\eg, $r=0.05$) to lose such information. As such, a model must learn to discriminate between color and texture information rather than shape, which we found often performs better. 

To perform the shapeless local patch discrimination, we first initialize the memory bank $\bm{V}$ to store features of all training images,
\begin{equation}
\bm{V} = [\mathbf{v}_1, \mathbf{v}_2,\cdots, \mathbf{v}_{N}]
\end{equation}
where $\mathbf{v_i}$ is the feature of the shapeless local patch $\mathbf{x}_i'$ from the $i$-th original image $\mathbf{x}_i$ (\ie, $\mathbf{v_i} = C_{SLPD}(F(\mathbf{x}_i')$) and $N$ is the total number of images. We randomly choose a square SLP $\mathbf{x}_i'$ out of the whole image (\eg, a random region cropped with $r=0.05$ of the whole area). Then, given an image $\mathbf{x}_j'$ in a minibatch , we compute the feature $\mathbf{f}_j =  C_{SLPD}(F(\mathbf{x}_j')$ minimize the contrastive loss~\cite{wu2018unsupervised} to discriminate the shapeless local patch, 
\begin{equation}
    \mathcal{L}_{SLPD} = - \log \frac{\exp ((\mathbf{v}_j)^\top \mathbf{f}_j / \tau)}{\sum_{k=1}^{N} \exp(({\mathbf{v}_k})^\top \mathbf{f}_j / \tau)},
\end{equation}
where the temperature parameter $\tau$ is the concentration level~\cite{hinton2015distilling}.

\subsection{Texture Discrimination (TD)}
\label{TD}
 Unlike the SLPD, texture discrimination (TD) uses the whole image to learn global texture patterns. Inspired by~\cite{lin2015bilinear,gatys2015neural}, we use a gram matrix (also called bilinear features) to obtain a texture representation for an image. Then, similar to the SLPD, we perform ID so items with similar textures embed nearby each other. First, we compute the feature map $\mathbf{g}_i = C_{TD}(F(x_i))$ of an input image $\mathbf{x_i}$ and a Gram matrix for texture representation~\cite{lin2015bilinear,gatys2015neural},
 \begin{equation}
     \mathbf{G}_i(j,k) = \mathbf{g}_i(j) \mathbf{g}_i(k)
 \end{equation}
 where $\mathbf{G}(j,k)$ is the inner product between the vectorized features of $j$-th and $k$-th channels in the feature map $\mathbf{g}_i$. In order to perform texture discrimination, we initialize the memory bank $\bm{T}$ to store texture representation of all training images.
 \begin{equation}
\bm{T} = [\mathbf{t}_1, \mathbf{t}_2,\cdots, \mathbf{t}_{N}]
\end{equation}
where $\mathbf{T}_i$ is the texture representation of $i$-th image (\ie,  $\mathbf{T}_i = \mathbf{G}_i(j,k)$). During training, similar to above, we compute the texture representation $\mathbf{G}_j$ of $x_j$ in a minibatch and minimize the contrastive loss~\cite{wu2018unsupervised} to discriminate texture representations between images, 
\begin{equation}
    \mathcal{L}_{TD} = - \log \frac{\exp ((\mathbf{t}_j)^\top \mathbf{G}_j / \tau)}{\sum_{k=1}^{N} \exp(({\mathbf{t}_k})^\top \mathbf{G}_j / \tau)},
\end{equation}


Finally, the overall learning objective for S-VAL is, 
\begin{equation}
    \hat{\theta} = \lambda_{rgb} \mathcal{L}_{rgb} +  \lambda_{SLPD} \mathcal{L}_{SLPD} + \lambda_{TD}\mathcal{L}_{TD}
\end{equation}
where $\lambda_{rgb}, \lambda_{SLPD}, \lambda_{TD}$ are the hyper-parameters for each loss. SLPD takes only shapeless local patches as input and TD takes the whole image to understand the global textures. Predicting the RGB histogram takes both types of input.

After updating the network parameters with each mini-batch $B$, we also update the memory features in the memory banks $\bm{V}$ and $\bm{T}$ with a momentum $\eta=0.5$ following~\cite{wu2018unsupervised}:
\begin{equation}
\begin{split}
\forall i \in B,
\quad
\mathbf{v}_i &=(1-\eta) 
\mathbf{v}_i+\eta C_{SLPD}(\mathbf{f}_i),
\\
\mathbf{t}_i &=(1-\eta) 
\mathbf{t}_i+\eta 
C_{TD}(\mathbf{f}_i),
\label{eq:memory_update}
\end{split}
\end{equation}


\section{Experiments}

Following Han~\etal~\cite{han2017learning}, we evaluate on the fashion compatibility and fill-in-the-blank (FITB) tasks as described below.
We denote the feature of an image $\mathbf{x}_i$ as $\mathbf{f}_{i}=F(\mathbf{x}_i)$.
\smallskip

\noindent\textbf{Fashion Compatibility.} In this task the goal is to discriminate between compatible and incompatible outfits. Following Han~\etal~\cite{han2017learning}, we report the area under a receiver operating characteristic curve (AUC) and average precision (AP) from the compatibility scores. Given N fashion items in an outfit, we compute the compatibility scores by computing the average pair-wise cosine similarities: $\frac{2}{N(N-1)}\sum_{i=0}^{N-1} \sum_{j=i+1}^{N-1} cos\_sim(\mathbf{f}_i, \mathbf{f}_j) $. 
\smallskip

\begin{table*}
\centering

\resizebox{0.8\textwidth}{!}{\begin{tabular}{|ll|c|c|c|c|}
\hline
 & &\multirow{2}{*}{Label?}  & \multicolumn{2}{c|}{Polyvore Outfits} & Capsule  \\
 \cline{4-6}
& Method & & Comp. AUC  &  FITB acc. & Comp. AP \\
\hline
\hline
\multirow{5}{*}{\shortstack{(a) With\\Label}} & Bi-LSTM~\cite{han2017learning} & Comp. &  0.65 & 39.7 & 18.4 \\
    & SiameseNet~\cite{vasileva2018learning} & Comp. & 0.81 & 52.9 & - \\
    & Type-Aware Network~\cite{vasileva2018learning} & Comp. &  0.86 & 55.3 & -\\
    & SCE-Net~\cite{tan2019learning} & Comp. & \textbf{0.91} & \textbf{61.6} & -  \\
    & Attribute Classifier & Attributes & 0.73  & 46.3  & 25.0 \\
    \hline
    \hline
\multirow{10}{*}{\shortstack{(b)\\ Self-sup.\\Baselines}} & ImageNet pre-trained & \xmark & 0.66 & 39.1  & 21.1  \\
    & Capsule Network (weakly-sup.)~\cite{hsiao2018creating} & \xmark  & - & -  & 19.9  \\
    & AutoEncoder~\cite{hinton1994autoencoders} & \xmark  & 0.58 & 34.0  & 19.8  \\
    & Colorization~\cite{zhang2016colorful} & \xmark  &  0.63 & 34.1 & 18.6  \\
    & Jigsaw~\cite{noroozi2016unsupervised} & \xmark  & 0.52 & 27.9  & 18.6  \\
    & Rotation~\cite{gidaris2018unsupervised} & \xmark  & 0.53 & 29.4  & 18.5  \\
    & ID \cite{wu2018unsupervised} w/ color distortion  & \xmark  & 0.57  & 30.8  & 18.9  \\
    & ID \cite{wu2018unsupervised} w/o color distortion & \xmark  & 0.74 & 45.9  & 23.3  \\
    & LA \cite{zhuang2019local} w/ color distortion  & \xmark & 0.56  & 30.4  & 19.1  \\
    & LA \cite{zhuang2019local} w/o color distortion & \xmark  & 0.74 & 46.3  & 24.0   \\
    \hline
\multirow{5}{*}{\shortstack{(c) S-VAL\\(Ours)}} & Predicting RGB histogram (RGB) & \xmark  & 0.77 & 47.2    &  23.3   \\
    & Shapeless Local Patch Disc. (SLPD) & \xmark  & 0.83 & 54.6 &  27.7   \\
    & Texture Disc (TD) & \xmark & 0.77 & 50.3    &  25.2   \\
    & RGB + SLPD & \xmark  & 0.83 & 55.4    &  27.7   \\
    & RGB + SLPD + TD  & \xmark  & \textbf{0.84} & \textbf{55.8}    &  \textbf{27.9}   \\
\hline
\end{tabular}}
\vspace{-2mm}
\caption{Comparison of (a) supervised models with compatibility or attribute labels and (b,c) unsupervised models on the Polyvore Outfits~\cite{vasileva2018learning} and Capsule~\cite{hsiao2018creating} datasets. \textit{All methods use ImageNet pre-trained weights and finetuned on Polyvore Outfits.} We report the performance of existing self-supervised learning baselines in (b) and our proposed approach in (c).}
\label{exp:tab_polyvore}
\vspace{-2mm}
\end{table*}

\noindent\textbf{Fill in the Black (FITB).} In this task the goal is to complete a partial outfit by selecting from a set of options. Similar to above, we compute the average similarity between each option and the partial outfit and select the one that gets the highest average compatibility. Performance is measured based on how often the choice was correct. 
\smallskip

\noindent\textbf{Fashion Retrieval.}
We also explore the fashion retrieval task. In fashion retrieval, the goal is to find the same item from a database given a query item that may be in a different view than those in the database. Similar to the fashion compatibility task, this task also needs some understanding of colors and textures, but shape also plays a factor since we are looking for exactly the same object. However, the shape of fashion items can still be changed significantly, as the items can appear in different poses, illumination, and camera angles. We report recall@k as our metric.
\smallskip

\noindent\textbf{Implementation details.} We use a ResNet-50~\cite{he2016deep} which is pre-trained on ImageNet~\cite{krizhevsky2012imagenet} for our feature extractor $F(\cdot)$ and all baselines. For each sub-tasks in Sec.~\ref{sec:method}, we attach the separate projection heads after the feature extractor. Following~\cite{chen2020simple}, these heads consist of two fully connected layers with ReLU activations followed by a $\ell_2$ normalization layer. All three self-supervised sub-tasks are trained jointly. We use each validation set to tune hyper-parameters for each sub-task and report averaged results over three runs. We randomly sample shapeless local patches with $r \in$ [0.05, 0.15] of the original image area.  We use a Adam optimizer optimizer~\cite{kingma2014adam} with a learning rate $5e^{-5}$. We train a model for 150 epochs and set the number of bins for each RGB channel as 10 and hyper-parameters $\lambda_{rgb}=1, \lambda_{SLPD}=1e^{-2}, \lambda_{TD}=1e^{-5}$ in Eq. 8 using the validation set~\cite{vasileva2018learning}. We set $\tau=0.07$ in Eqs. 4 and 7 following~\cite{wu2018unsupervised}.

We also provide the following self-supervised baselines for comparison:  AutoEncoder~\cite{hinton1994autoencoders}, colorization~\cite{zhang2016colorful}, sovling jigsaw puzzles~\cite{noroozi2016unsupervised}, predicting rotation~\cite{gidaris2018unsupervised}, Instance Discrimination (ID)~\cite{wu2018unsupervised}, and  Local Aggregation~\cite{zhuang2019local}. Please note that all methods finetune the same ResNet-50 initialized with ImageNet pretrained weights as our approach.


\subsection{Datasets}

\noindent\textbf{Polyvore Outfits~\cite{vasileva2018learning}} has 53,306 outfits from 204K images for training, 10K outfits from 47K images for testing and 5K outfits from 25K images for validation. We use the provided fashion compatibility and FITB questions, where items in ground truth outfits were replaced with random items of the same type for fashion compatibility, or 3 random items of the same type were selected as incorrect answers for FITB (resulting in 4 choices).  We also use the Polyvore-D split that contains 71K images.  In this split no item that appears in the training outfits also appears in the testing outfits.
\smallskip


\noindent\textbf{Capsule Wardrobe~\cite{hsiao2018creating}} contains 15K fashion compatibility questions from 6K images, which are all used for testing.  We train on the Polyvore Outfits dataset when evaluating on Capsule Wardrobe.
\smallskip

\noindent\textbf{Fashion-Gen~\cite{rostamzadeh2018fashion}} has 260K images of luxury fashion items 
with descriptions. We only train on this dataset and evaluate on Polyvore Outfits since no outfit information is publicly available.
\smallskip

\noindent\textbf{In-Shop Clothing Retrieval benchmark in DeepFashion~\cite{liu2016deepfashion}} contains 52K images of 8K clothing items from web data containing large poses and scale variations. This benchmark splits its test data into a query and gallery set, where no items in either of these sets are shared with those seen during training.

\subsection{Unsupervised Evaluation Results}
Table \ref{exp:tab_polyvore} shows results on the Polyvore~\cite{vasileva2018learning} and Capsule Wardrobe test set~\cite{hsiao2018creating}. In Table \ref{exp:tab_polyvore}(a), we report the performance of supervised models with trained compatibility labels or attribute labels in Polyvore as a reference. In Table \ref{exp:tab_polyvore}(b), we report the performance of the self-supervised learning baselines fine-tuned on Polyvore from the ImageNet pre-trained model. We see that existing self-supervised learning methods including reconstruction based methods~\cite{zhang2016colorful,hinton1994autoencoders} and handcrafted sub-tasks~\cite{gidaris2018unsupervised,noroozi2016unsupervised} actually harm performance compared to the ImageNet pre-trained model. We also observe that ID and Local Aggregation with color distortion underperform the ImageNet pre-trained model. When we remove the color distortion augmentation in their methods, these methods outperform the ImageNet pre-trained model. These results suggest that directly applying the existing self-supervised learning methods does not help on the fashion compatibility task. From now on, we remove the color distortion augmentation in ID and Local Aggregation for all other comparisons.

We show the performance of our method in Table \ref{exp:tab_polyvore}(c) including an ablation analysis. We observe that each sub-task predicting RGB histograms (Sec.~\ref{sec:rgb}), shapeless local patch discrimination (Sec.~\ref{sec:slpd}), and texture discrimination (Sec.~\ref{TD}), improves the performance over the ImageNet pre-trained network. Combing all three components gets the best performance, resulting in a 9.5-10 points improvement on Polyvore Outfits over prior SSL baselines, and 4 points better on Capsule Wardrobes. In addition to outperforming the SSL baselines, our full model without any labels outperforms simple the Simaese Network trained with compatibility labels, as well as Bi-LSTM~\cite{han2017learning}, while also being comparable to the fully-supervised Type-Aware Network.

\subsection{Additional Analysis}

\noindent\textbf{Polyvore-D and Cross Dataset Evaluation.} Table~\ref{exp:tab_others_1} shows the comparison on Polyvore-D containing three times fewer training images than Polyvore Outfits. Table~\ref{exp:tab_others_2} explores a cross dataset evaluation scenario, where a model is trained on Fashion-Gen but evaluated on Polyvore Outfits. In both cases, our approach outperforms the best SSL baseline, Local Aggregation, by 8-9 points on both tasks.

\begin{table}[t]
\centering
\small
\vspace{-1.5mm}
\centering
\resizebox{0.37\textwidth}{!}{\begin{tabular}{|l|c|c|}
\hline
 & \multicolumn{2}{c|}{Polyvore-D} \\
\hline
    Method & Comp. AUC & FITB acc.   \\
\hline
\hline
    ID~\cite{wu2018unsupervised}& 0.69 & 43.2 \\
    LA~\cite{zhuang2019local} & 0.73 & 46.2 \\
    \hline
    RGB & 0.74 & 45.7  \\ 
    RGB+SLPD & \textbf{0.81} & 53.9 \\
    RGB+SLPD+TD & \textbf{0.81} & \textbf{54.3} \\
\hline
\end{tabular}}
\caption{Fashion compatibility evaluation on the Polyvore-D Split. The Polyvore-D split containg less training data than Polyvore. Our method outperforms the baselines.}
\label{exp:tab_others_1}
\end{table}

\begin{table}[t]
\centering
\resizebox{0.37\textwidth}{!}{\begin{tabular}{|l|c|c|}
\hline
 & \multicolumn{2}{c|}{Fashion-Gen $\rightarrow$ Polyvore} \\
 \hline
    Method & Comp. AUC & FITB acc.   \\
\hline
\hline
    ID~\cite{wu2018unsupervised} & 0.71 & 45.5 \\
    LA~\cite{zhuang2019local} & 0.73 & 46.5 \\
    \hline
    RGB & 0.76 & 48.1  \\ 
    RGB+SLPD & 0.80 & 52.9 \\
    RGB+SLPD+TD & \textbf{0.81} & \textbf{53.3} \\

\hline
\end{tabular}}
\caption{Cross dataset evaluation on the fashion compatibility task. We train a model on the Fashion-Gen dataset and test it on the Polyvore dataset. We report the number of self-supervised learning baselines and ours. Our method is able to generalize across different datasets.}
\label{exp:tab_others_2}
\vspace{-4mm}
\end{table}


\smallskip

\begin{figure*}[t!]
	\centering
    \begin{subfigure}[t]{0.24\textwidth}
        \includegraphics[width=1.0\linewidth, height=2.7cm]{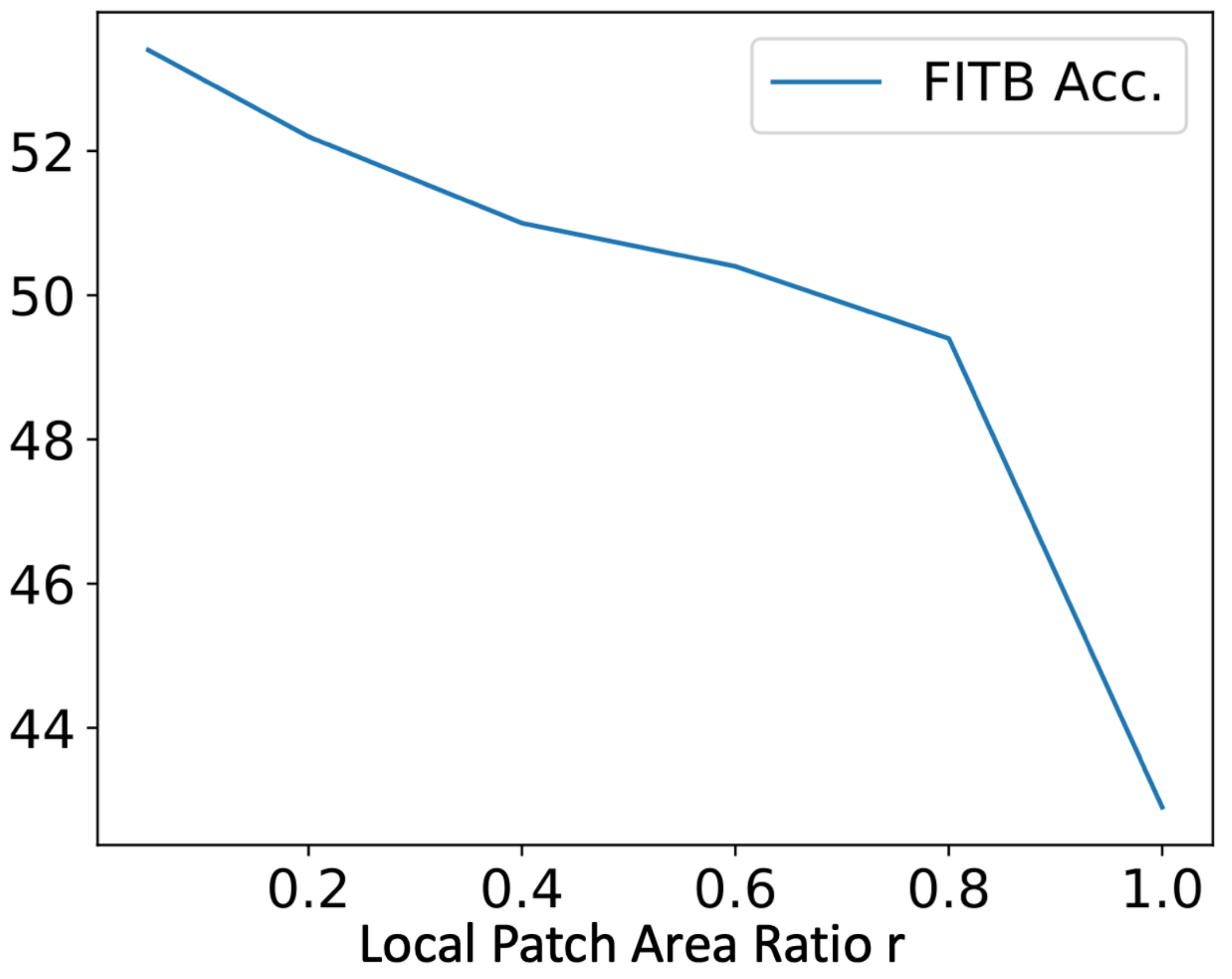}        \caption{}
    \end{subfigure}
    ~
	\begin{subfigure}[t]{0.24\textwidth}
        \centering
        \includegraphics[width=0.97\linewidth, height=2.7cm]{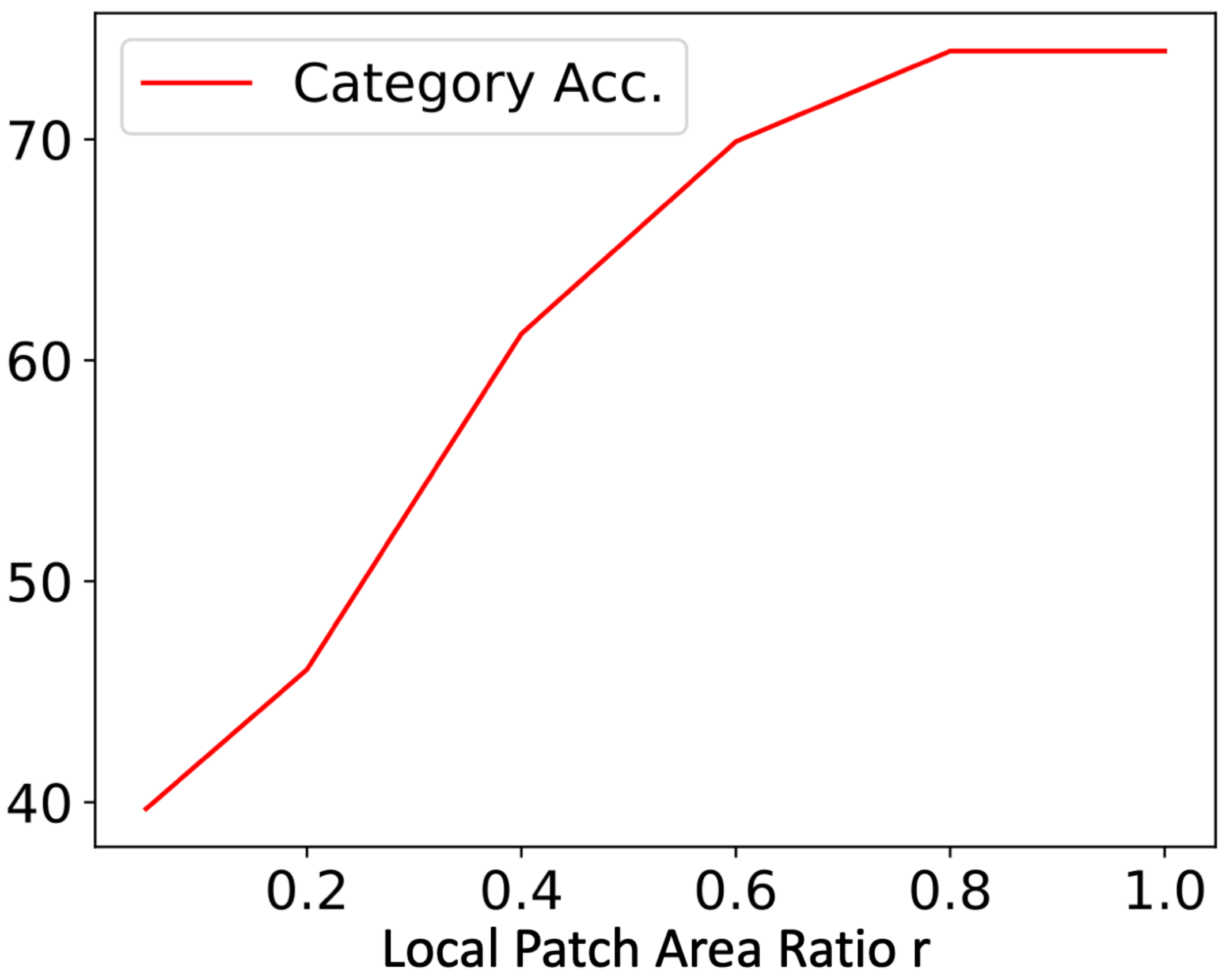}
        \caption{}
    \end{subfigure}
    ~
    \begin{subfigure}[t]{0.23\textwidth}
        \centering
        \includegraphics[width=1.0\linewidth, height=2.8cm]{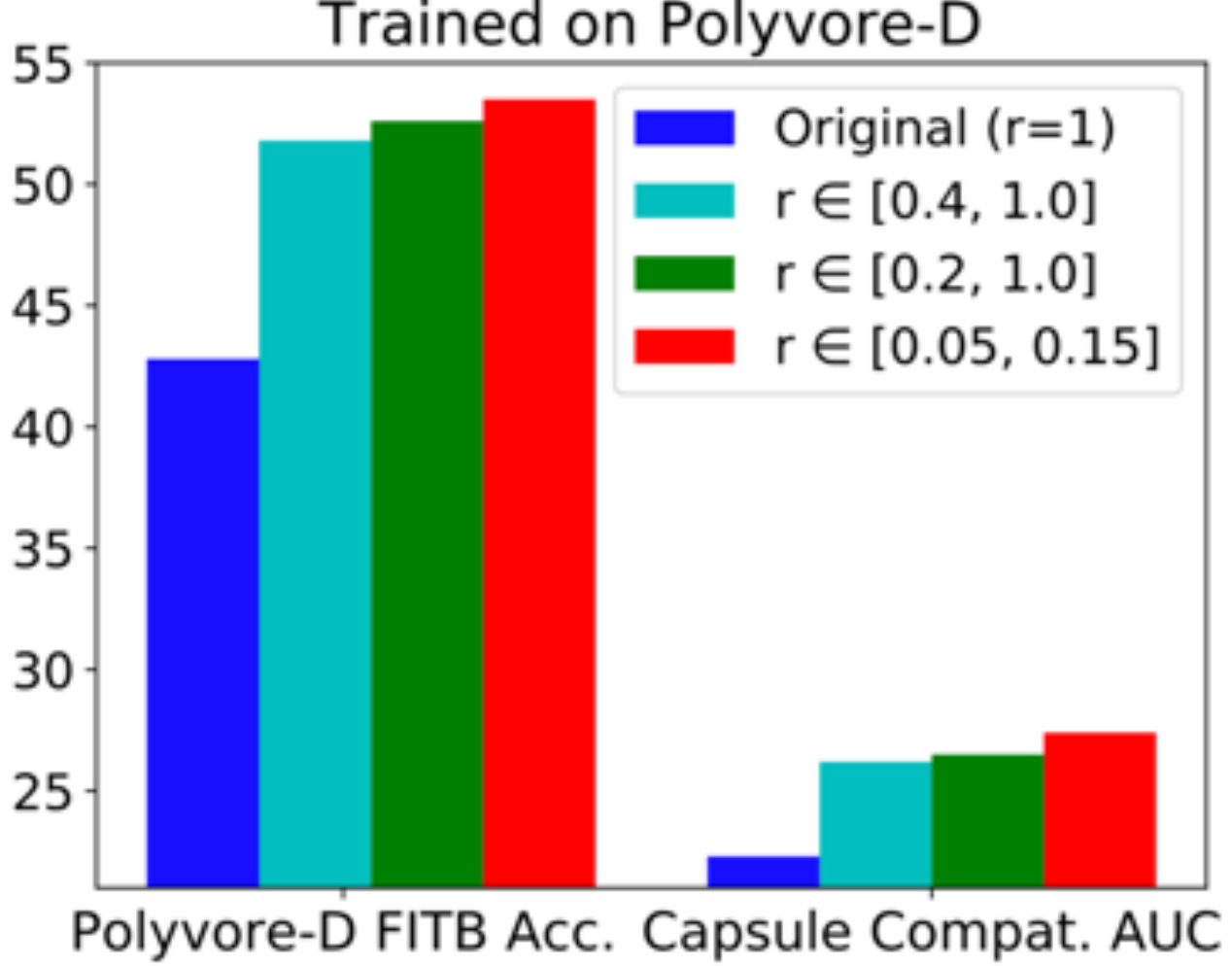}
        \caption{}
    \end{subfigure}
    ~
    \begin{subfigure}[t]{0.23\textwidth}
        \centering
        \includegraphics[width=1.0\linewidth, height=2.8cm]{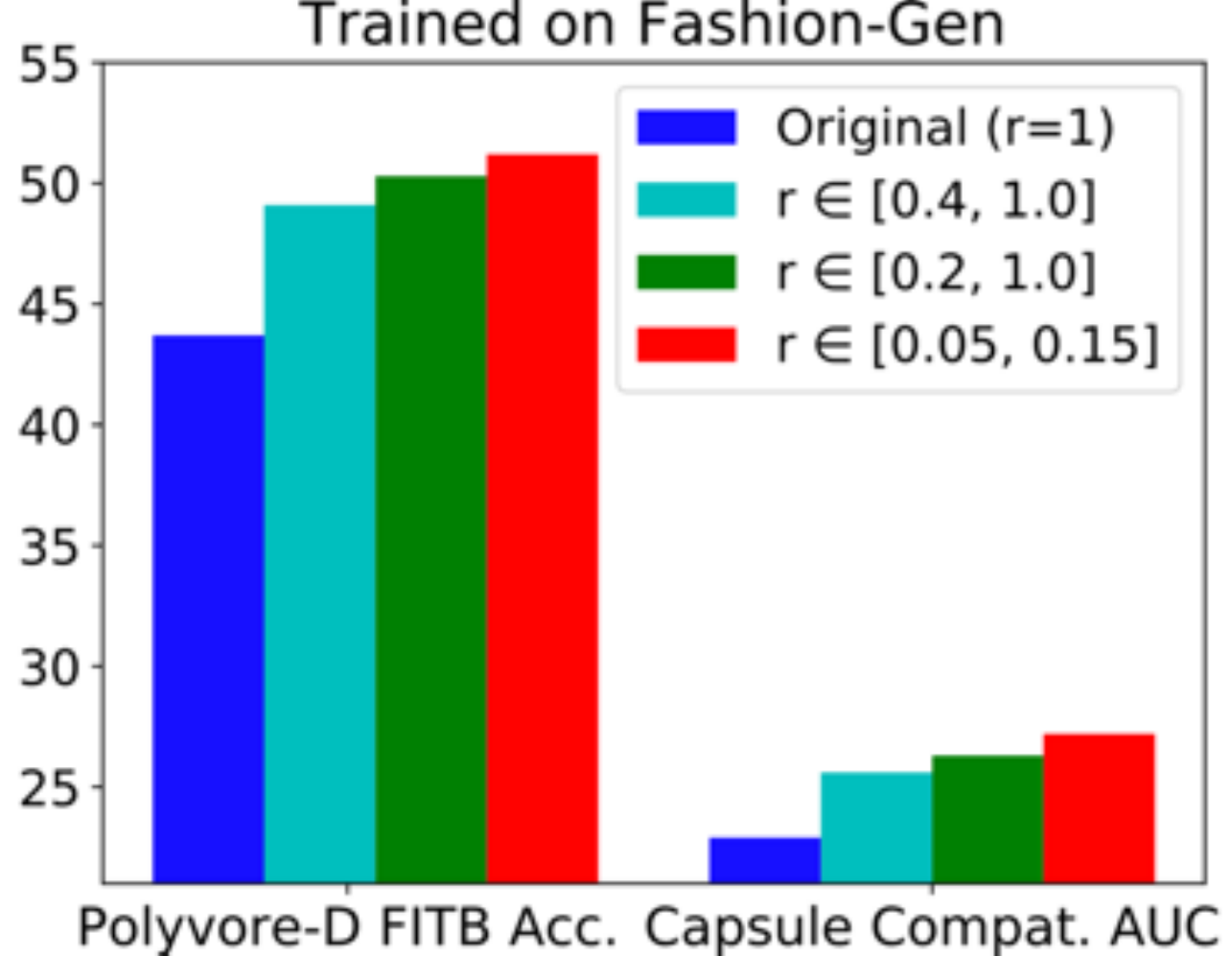}
        \caption{}
    \end{subfigure}
    \vspace{-2mm}
	\caption{Ablation study on the effect of local patch area ratio $r$ on Polyvore-D. In (a,b), we report the performances of the task of fashion compatibility and object recognition according to the different area ratios of the local patch. In (c,d), we provide the comparisons on original input size $r$=1 and random cropping with different ratios in the specified range during training. These results show that using smaller patches performs better while generating shape invariant features than using larger patches.}
	\label{fig:fig_ablation}
\end{figure*}

\begin{figure*}[t!]
	\centering
	\begin{subfigure}[t]{0.49\textwidth}
        \centering
        \includegraphics[width=0.95\linewidth]{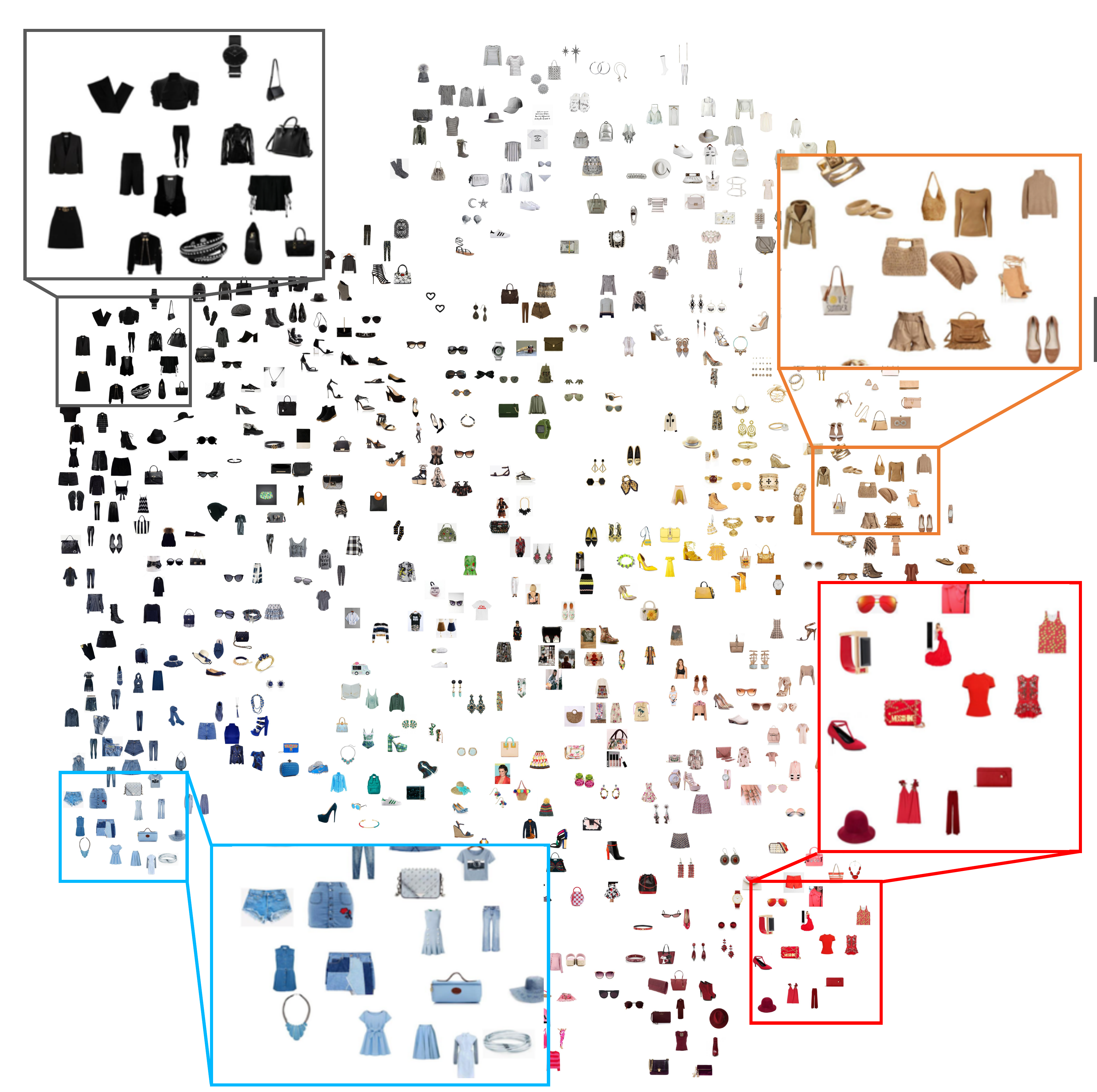}
        \caption{S-VAL (ours)}
    \end{subfigure}
    ~
	\begin{subfigure}[t]{0.49\textwidth}
        \centering
        \includegraphics[width=0.95\linewidth]{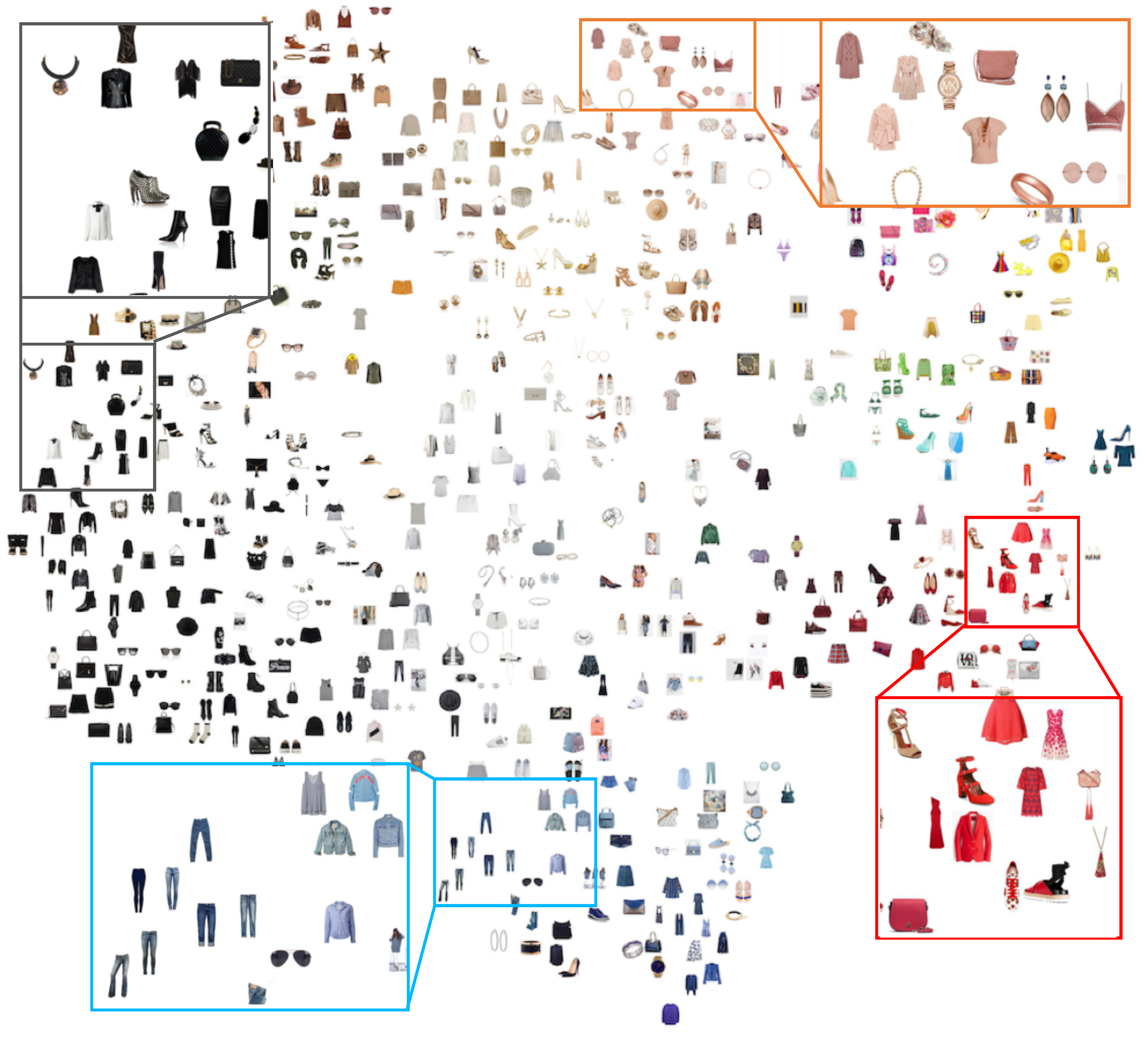}
        \caption{Siamese Network (supervised)}
    \end{subfigure}

    \begin{subfigure}[t]{0.49\textwidth}
        \centering
        \includegraphics[width=0.95\linewidth]{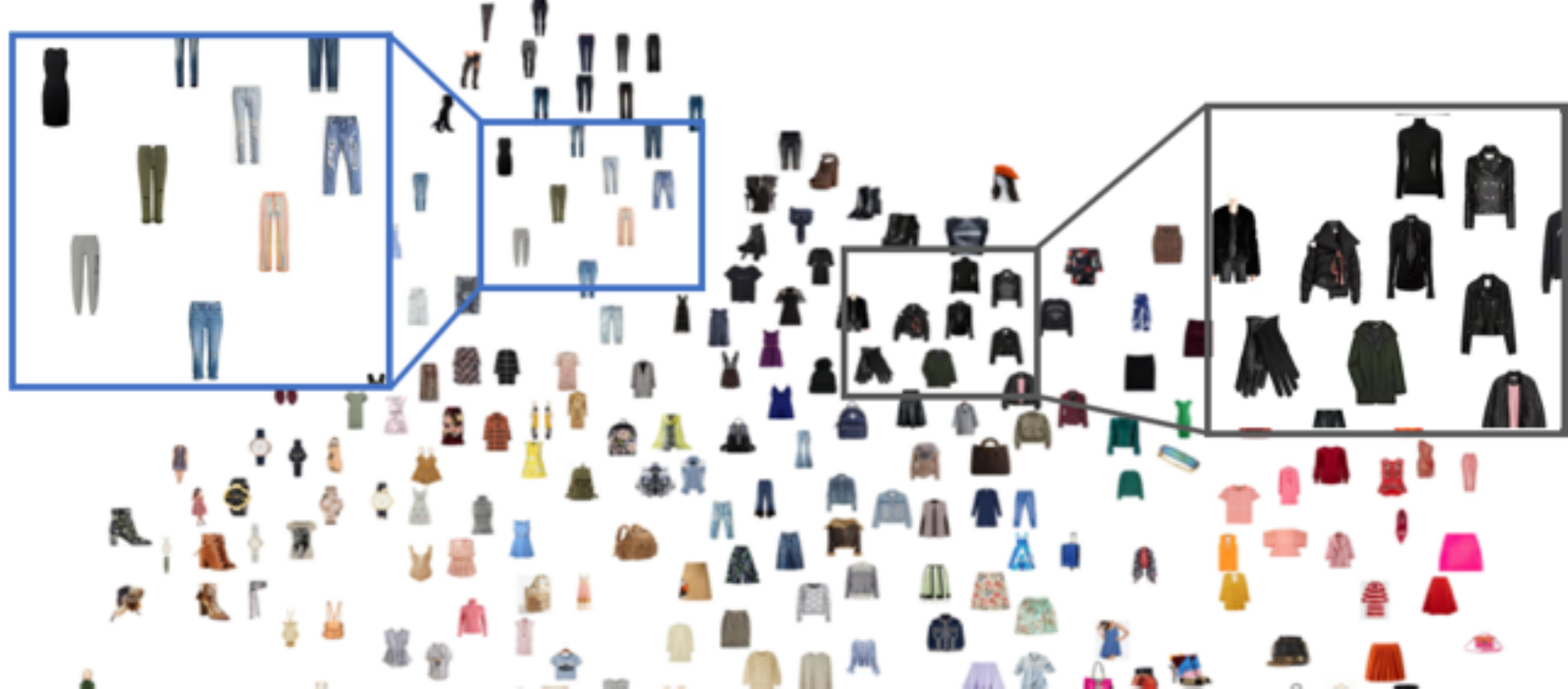}
        \caption{ImageNet pre-trained Network}
    \end{subfigure}
    ~
	\begin{subfigure}[t]{0.49\textwidth}
        \centering
        \includegraphics[width=0.95\linewidth]{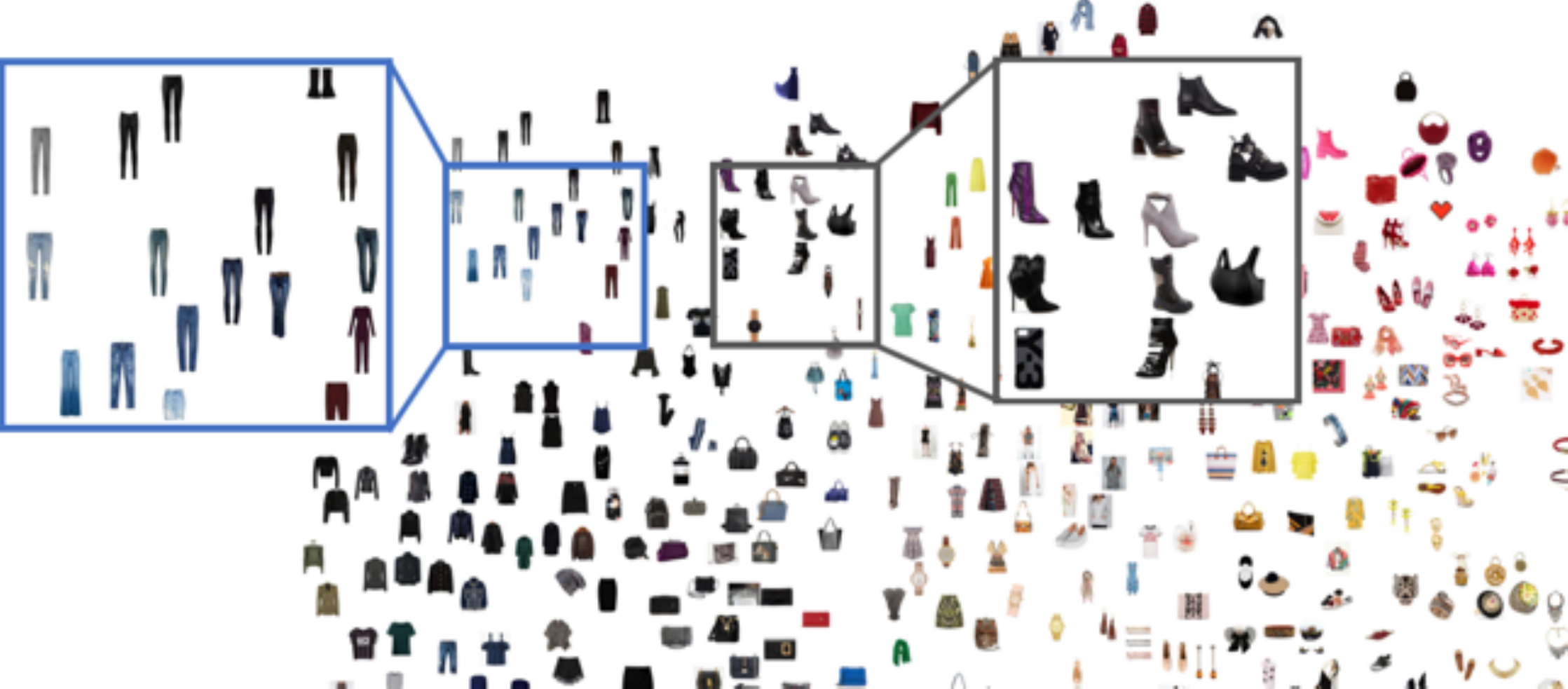}
        \caption{ID}
    \end{subfigure}
	\caption{t-SNE visualizations. Similar to (b) the supervised model,  (a) our unsupervised model learns a similar embedding which embeds items with similar visual attributes (\eg, colors and texture) nearby regardless of object categories. While the ImageNet pre-trained network and ID generate features biased to object shapes, items with different visual attribute can be embedded nearby.
    }
	\label{fig:fig_tsne}
\end{figure*}

\noindent\textbf{Ablation Study on Patch Area Ratio.} 
In this section, we analyze how the different area ratios affect the performance on both fashion compatibility  and object recognition (denoted by ``Category Acc'') in Fig.~\ref{fig:fig_ablation}. We measure the object recognition accuracy with a kNN classifier~\cite{wu2018unsupervised} on image features. In Fig.~\ref{fig:fig_ablation}(a), we report the FITB accuracy using different local patch sizes. It is clear that using a small local patch improves performance considerably over using a large local patch. Fig.~\ref{fig:fig_ablation}(b) reports category recognition accuracy, which appears to have an inverse relationship with~\ref{fig:fig_ablation}(a), demonstrating that addressing fashion compatibility requires different methods than typically used in prior work in SSL that mainly investigated methods for object recognition. Finally in Fig.~\ref{fig:fig_ablation}(c, d), we compare models trained with ID using different area ratios $r$: original image only (\ie, $r=1.0$) and different random cropping ratios of $r \in [0.4, 1.0], [0.2, 1.0],[0.05, 0.15]$. We see that using larger patches harms the performance compared to using smaller patches only. These results also suggest that the performance gain mostly comes from the small patches. Thus, training with very small local patches losing shape clues is a key component in SSL for fashion compatibility.
\smallskip

\noindent\textbf{Visualization.}
Figure \ref{fig:fig_tsne} shows t-SNE visualizations~\cite{maaten2008visualizing} of features on Polyvore from each model. We also confirm the observation of~\cite{plummerSimilarityExplanations2020} that the Siamese network trained on compatibility labels embeds similar color or texture items nearby ignoring fashion item categories (the third row in Table~\ref{exp:tab_polyvore}(a)). By comparing the Fig. \ref{exp:tab_polyvore}(a, b), our model produces a very similar feature distribution as the Siamese network. Both models tend to cluster similar items nearby in terms of colors and texture regardless of object categories. However, Fig.~\ref{fig:fig_tsne}(c) and (d) cluster items based on shape, so that items with different attributes from the same object class are embedded nearby, which could be harmful to the fashion compatibility task as discussed earlier.

\begin{figure}[t]
	\centering
    \includegraphics[width=\linewidth]{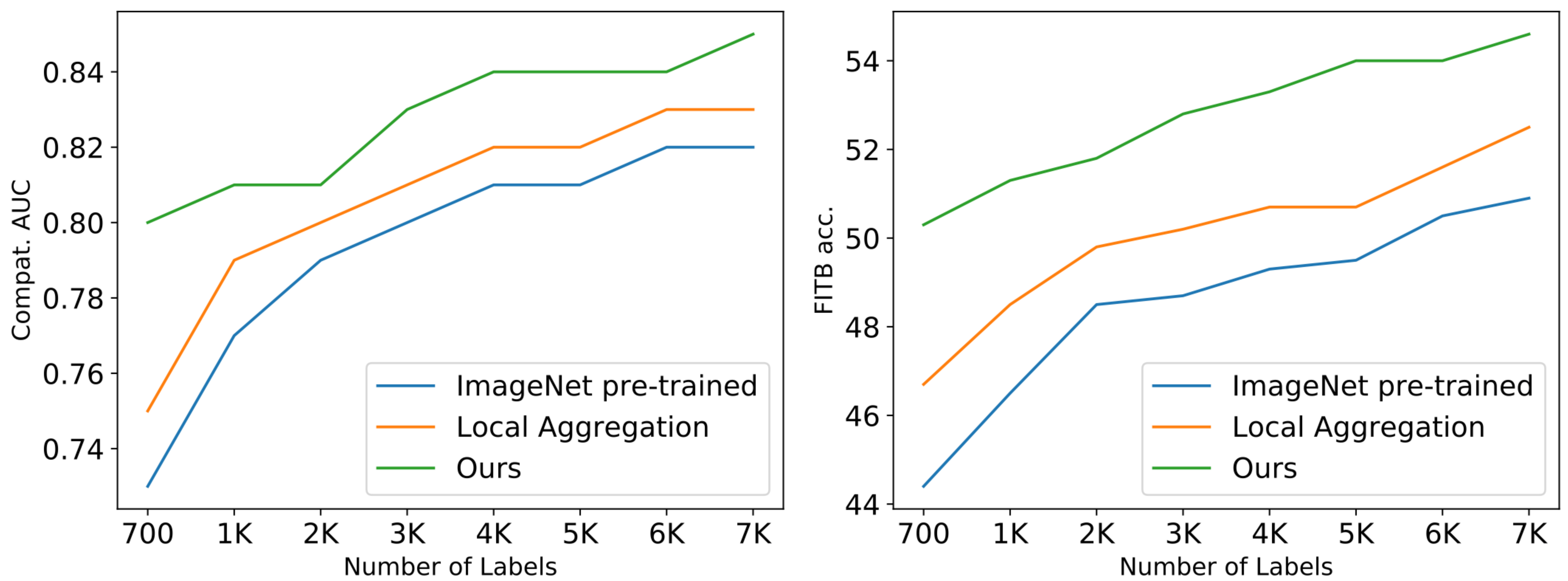}
    \caption{Comparison under linear classification protocol with fashion compatibility labels. ``Ours'' denotes our full method, RGB + SLPD + TD.
    }
	\label{fig:fig_linear}
	\vspace{-2mm}
\end{figure}

\noindent\textbf{Linear Classification Protocol.} We evaluate our method on a linear classification protocol~\cite{wu2018unsupervised,he2019momentum,chen2020simple}. In this evaluation, we use fixed image features $\mathbf{f}\in\mathbb{R}^{2048}$ and train only a linear classifier $\mathbf{W}\in\mathbb{R}^{2048\times64}$ on compatibility labels using triplet loss. To effectively evaluate the features learned from SSLs, we report performance when different numbers of training labels are available in Fig.~\ref{fig:fig_linear}. We compare ours with the ImageNet pre-trained network and Local Aggregation, which is the best performing self-supervised baseline. We observe that our method consistently outperforms other baselines and the benefit of our method is more significant when there are fewer labels.

\noindent\textbf{Fashion Retrieval Evaluation.} In Table~\ref{exp:tab_fashion_retrieval}, we report the accuracy of recall@k of the DeepFashion Inshop retrieval task~\cite{liu2016deepfashion}. We start with an ImageNet pre-trained model and use SSL methods without labels. As a reference, we show the accuracy of a fully supervised model with 200K pair annotations in the first row of Table~\ref{exp:tab_fashion_retrieval}. We use a standard triplet loss to train the supervised model similar to a Siamese Network in~\cite{vasileva2018learning}. Different from the fashion compatibility task, the ID with color distortion augmentations improves performance compared to the ImageNet pre-trained model. By removing the color distortion augmentations, ID further improves the performance. As expected, this result shows that shape information is helpful to learn useful features for the fashion retrieval task. Then we perform each of the components in S-VAL. In this task, predicting RGB histogram does not help much. This could be because predicting RGB histogram does not consider item shapes and enforce a model to produce invariant features to object shapes. As learning shape is also important to retrieve the same category item in this task, predicting RGB histogram is not desirable. We see that SLPD and TD outperform ID by a large margin by learning color patterns in a local patch and global texture patterns from TD. These results suggest that directly applying the existing method on any downstream task is not the best option. We argue that self-supervised learning methods should consider the characteristics of a downstream task. 

\begin{table}[t]
\centering
\small
\vspace{-1.5mm}
\centering
\resizebox{0.48\textwidth}{!}{\begin{tabular}{|l|c|c|c|}
\hline
 & \multicolumn{3}{c|}{DeepFashion In-shop Retrieval~\cite{liu2016deepfashion}} \\
\hline
    Method & Recall@1 & Recall@5 & Recall@10   \\
\hline
\hline
    Triplet (supervised)  & 63.6 & 85.4 & 89.3  \\
    \hline
    \hline
    ImageNet pre-trained & 16.8 & 36.4 & 42.5\\
    ID w/ color distortion & 25.5 & 51.5 & 60.1 \\
    ID w/o color distortion & 30.0 & 56.1 & 67.6 \\
    \hline
    RGB & 25.0 & 48.5 & 55.6 \\ 
    SLPD & 39.9 & 64.6 & 70.6 \\
    TD & 29.7 & 54.9 & 64.2 \\
    SLPD+TD & \textbf{46.5} &74.8 &  \textbf{81.3} \\
    SLPD+TD+RGB & 46.2 &  \textbf{75.0} & 81.2\\
\hline
\end{tabular}}
\caption{Evaluation on the in-shop fashion retrieval task~\cite{liu2016deepfashion}. The top row reports the accuracy of the supervised model with a triplet loss for reference. We report the unsupervised fashion retrieval accuracy using Recall@k.  }
\label{exp:tab_fashion_retrieval}
\vspace{-4mm}
\end{table}

\section{Conclusion}
While prior self-supervised learning approaches have been successful, their downstream task is mostly related to object recognition which focuses on learning object shape variant and color invariant features. In this paper, we explore self-supervised methods for the fashion compatibility and retrieval task, where colors and texture are important.  We propose a new Self-supervised Tasks for Visual Attribute Learning (S-VAL) which learns colors and texture patterns while generating shape-invariant features. Our method is built upon an observation that similar color or texture items are more likely compatible, but it is possible that different color items can be matched. We also show that prior work in self-supervised learning often fails to generalize to computer vision tasks that require a model that learns visual cues other than object shape. On the fashion compatibility task, S-VAL outperforms prior self-supervised learning approaches by 9.5-16\% and by 16.5\% in the fashion retrieval task. Notably, our approach obtained similar performance to some fully-supervised methods from prior work of fashion compatibility despite the fact our approach does not use any labels. We hope that our work will inspire research in self-supervised learning in additional application areas, as well as provide valuable insights to improve fashion recommendation systems in future work.

\section{Acknowledgements}
 This work was supported by NSF and the DARPA LwLL program.

{\small
\bibliographystyle{ieee_fullname}
\bibliography{egbib}
}

\end{document}